%% file: main.tex
\documentclass[10pt,twocolumn,letterpaper]{article}

\usepackage{cvpr}              


\definecolor{cvprblue}{rgb}{0.21,0.49,0.74}
\usepackage[pagebackref,breaklinks,colorlinks,allcolors=cvprblue]{hyperref}

\def\paperID{*****} 
\def\confName{CVPR}
\def\confYear{2025}

\title{SuperAD: A Training-free Anomaly Classification and Segmentation Method for CVPR 2025 VAND 3.0 Workshop Challenge Track 1: Adapt \& Detect}

\author{Huaiyuan Zhang$^{1}$
\and
Hang Chen$^{1}$
\and
Yu Cheng$^{1}$
\and 
Shunyi Wu$^{1}$
\and
Linghao Sun$^{1}$
\and
Linao Han$^{1}$
\and 
Zeyu Shi$^{2}$
\and 
Lei Qi$^{1}$\thanks{Corresponding Author}\footnotemark[1]
\and
$^{1}$School of Computer Science and Engineering, Southeast University, China\\
$^{2}$School of Computer Science and Engineering, Nanjing University of Science and Technology, China\\
{\tt\small \{zhang\_hy, hangchen, chengyu, shunyiwu, linghaosun, linaohan, qilei\}@seu.edu.cn}\\
{\tt\small shizeyu@njust.edu.cn}\\
}

\begin{document}
\maketitle
\input{sec/0_abstract}    
\input{sec/1_intro}

\input{sec/2_methodology}

\input{sec/3_results}

\input{sec/4_discussion}
\input{sec/5_conclusion}
{
    \small
    \bibliographystyle{ieeenat_fullname}
    \bibliography{main}
}


\end{document}

%% file: sec/0_abstract.tex
\begin{abstract}
In this technical report, we present our solution to the CVPR 2025 Visual Anomaly and Novelty Detection (VAND) 3.0 Workshop Challenge Track 1: \textit{Adapt \& Detect: Robust Anomaly Detection in Real-World Applications}. In real-world industrial anomaly detection, it is crucial to accurately identify anomalies with physical complexity, such as transparent or reflective surfaces, occlusions, and low-contrast contaminations. The recently proposed MVTec AD 2 dataset significantly narrows the gap between publicly available benchmarks and anomalies found in real-world industrial environments. To address the challenges posed by this dataset—such as complex and varying lighting conditions and real anomalies with large scale differences—we propose a fully training-free anomaly detection and segmentation method based on feature extraction using the DINOv2 model named \textbf{SuperAD}. Our method carefully selects a small number of normal reference images and constructs a memory bank by leveraging the strong representational power of DINOv2. Anomalies are then segmented by performing nearest neighbor matching between test image features and the memory bank. Our method achieves competitive results on both test sets of the MVTec AD 2 dataset.
\end{abstract}

%% file: sec/1_intro.tex
\section{Introduction}
\label{sec:intro}

\subsection{Background}
Unsupervised anomaly detection and localization has emerged as a key technology in computer vision, with wide-ranging applications in real-world scenarios such as industrial quality inspection and autonomous driving. The central challenge of this task lies in training models solely on normal samples, while enabling accurate identification and precise localization of previously unseen defects during testing. In recent years, fueled by the advancement of deep learning, numerous methods have achieved remarkable progress on mainstream benchmark datasets such as MVTec AD and VisA. However, as model performance on these datasets approaches saturation---for instance, segmentation AU-PRO scores of some algorithms on MVTec AD have surpassed 97\%---their limitations have become increasingly apparent:

\begin{enumerate}
    \item \textbf{Limited Scene Diversity}: Existing datasets primarily focus on objects with clear textures and simple structures, lacking coverage of complex industrial scenarios involving transparent or reflective surfaces (e.g., glassware, metal products), as well as bulk, overlapping items (e.g., granular materials).
    
    \item \textbf{Idealized Defect Types}: Most defects in current datasets are large, centrally located anomalies, overlooking real-world industrial defects such as edge anomalies, subpixel-level scratches (e.g., hairline cracks), and low-contrast contaminations (e.g., transparent foreign objects).
    
    \item \textbf{Insufficient Environmental Robustness}: These datasets often ignore variations in lighting conditions (e.g., dark-field, backlighting, overexposure), resulting in models with limited generalization ability when deployed across different devices or under varying environments.
\end{enumerate}

\subsection{Challenge Description}
As a next-generation benchmark for industrial anomaly detection, the MVTec AD 2 dataset~\cite{heckler2025mvtec} systematically addresses the aforementioned limitations by introducing eight complex and diverse scenarios. Its core features include:

\begin{itemize}
    \item \textbf{Simulation of Physical Complexity}:
    \begin{itemize}
        \item \textit{Transparent and reflective surfaces}: Categories such as \texttt{Vial} feature liquid refraction artifacts, and \texttt{Sheet Metal} includes mirror-like reflections, both of which challenge the model's ability to reason about light propagation.
        \item \textit{Bulk and overlapping objects}: Examples like \texttt{Wallplugs} and \texttt{Walnuts} involve random occlusions and truncated boundaries between objects, requiring semantic-level understanding.
        \item \textit{High intra-class variability in normal samples}: Categories such as \texttt{Fabric} (with diverse textures) and \texttt{Can} (with geometric pattern deformations) demand models to learn tight boundaries of the normal data manifold.
    \end{itemize}

    \item \textbf{Verification of Detection Limits}:
    \begin{itemize}
        \item \textit{Tiny objects and boundary anomalies}: Plastic contaminants occupying less than 0.1\% of the image area in the \texttt{Rice} category, and missing regions at the image boundaries in \texttt{Wallplugs} pose significant challenges to the model's perceptual capability at high resolution.
        \item \textit{Implicit structural consistency}: \texttt{Fruit Jelly} require the model to assess the plausibility of ingredient distributions, despite the absence of explicit logical constraints.
    \end{itemize}

    \item \textbf{Cross-Domain Generalization Evaluation}:
    Each scene includes at least four lighting conditions (regular, underexposed, overexposed, and additional light sources), simulating distribution shifts caused by device variations in real-world production environments. This enables systematic evaluation of model robustness.
\end{itemize}

On the MVTec AD 2 dataset, mainstream methods exhibit clear performance bottlenecks:

\textbf{Limited localization capability}: EfficientAD~\cite{batzner2024efficientad} and PatchCore~\cite{roth2022towards} achieve average AU-PRO\textsubscript{0.05} scores~\cite{heckler2025mvtec} of only 58.7\% and 53.8\%, respectively. In highly complex scenarios such as \texttt{Can} and \texttt{Rice}, their performance drops below 30\%.
    
\textbf{Poor robustness}: MSFlow~\cite{zhou2024msflow} shows a significant performance degradation of 51.1\% in AU-PRO\textsubscript{0.05} on the mixed lighting test set \textit{TEST\textsubscript{priv,mix}} compared to the standard test set \textit{TEST}\textsubscript{priv}, highlighting its sensitivity to environmental variations.

In summary, this paper targets the key challenges highlighted in the MVTec AD 2 dataset and aims to:
\begin{itemize}
    \item Develop a novel and highly robust model capable of accurately detecting subtle defects within highly variant normal samples;
    \item Enhance localization performance on small targets and structurally complex objects, such as transparent or overlapping instances;
    \item Strengthen the model's generalization ability under varying illumination conditions, thereby improving its practical applicability in industrial deployment.
\end{itemize}

%% file: sec/2_methodology.tex
\section{Methodology}
\label{sec:methodology}

\subsection{Model Design}
\subsubsection{Approach}
In recent years, memory bank-based methods have demonstrated remarkable performance on various anomaly detection and segmentation benchmarks. PatchCore~\cite{roth2022towards} introduces a greedy coreset selection mechanism to construct a compact memory bank, significantly reducing the number of features while preserving the overall distribution of the original feature space. This design effectively balances detection accuracy and retrieval efficiency. DMAD~\cite{hu2024dmad} unifies two commonly encountered scenarios in industrial settings: one with only normal samples available and the other with a limited number of labeled anomalies. It achieves this by constructing a normal memory bank and an expandable anomaly memory bank, which store features of normal and observed anomalous patterns, respectively, thereby improving adaptability to real-world complexities.

It is worth noting that the success of these memory-based approaches heavily depends on the quality of feature extraction. As a result, most methods rely on powerful pre-trained visual backbones such as WideResNet~\cite{zagoruyko2016wide} or Vision Transformer (ViT)~\cite{dosovitskiy2020image}. Prior studies reveal that different layers in deep models capture distinct types of information: shallow layers tend to focus on local high-frequency details (e.g., textures, edges), whereas deeper layers encode more abstract semantic information. Therefore, combining both shallow and deep features enables the model to capture both global structure and local details, which is crucial for accurate anomaly detection.

For instance, PaDiM~\cite{defard2021padim} employs a pre-trained ResNet to extract patch-wise features from four different layers and models the distribution of normal features at each spatial location using a multivariate Gaussian distribution, parameterized by the mean and covariance. APRIL-GAN~\cite{chen2023april}, on the other hand, leverages CLIP's powerful multimodal alignment capabilities. It extracts four-layer features from both the test image and its corresponding normal reference image, performs layer-wise matching, and averages the results to localize anomalous regions.

Fortunately, recent advances in self-supervised learning, such as DINOv2~\cite{oquab2023dinov2}, have shown strong capability in capturing rich semantic information for visual tasks. Building upon this, our method constructs a class-specific normal feature memory bank for each category in the MVTec AD 2 dataset. During inference, the features from various regions of an input image are matched against those stored in the memory bank to detect anomalies. We adopt the powerful DINOv2 backbone to extract multi-level features, aiming to achieve accurate and fine-grained anomaly segmentation.

\begin{figure*}[t]
  \centering
  \includegraphics[width=1.0\textwidth]{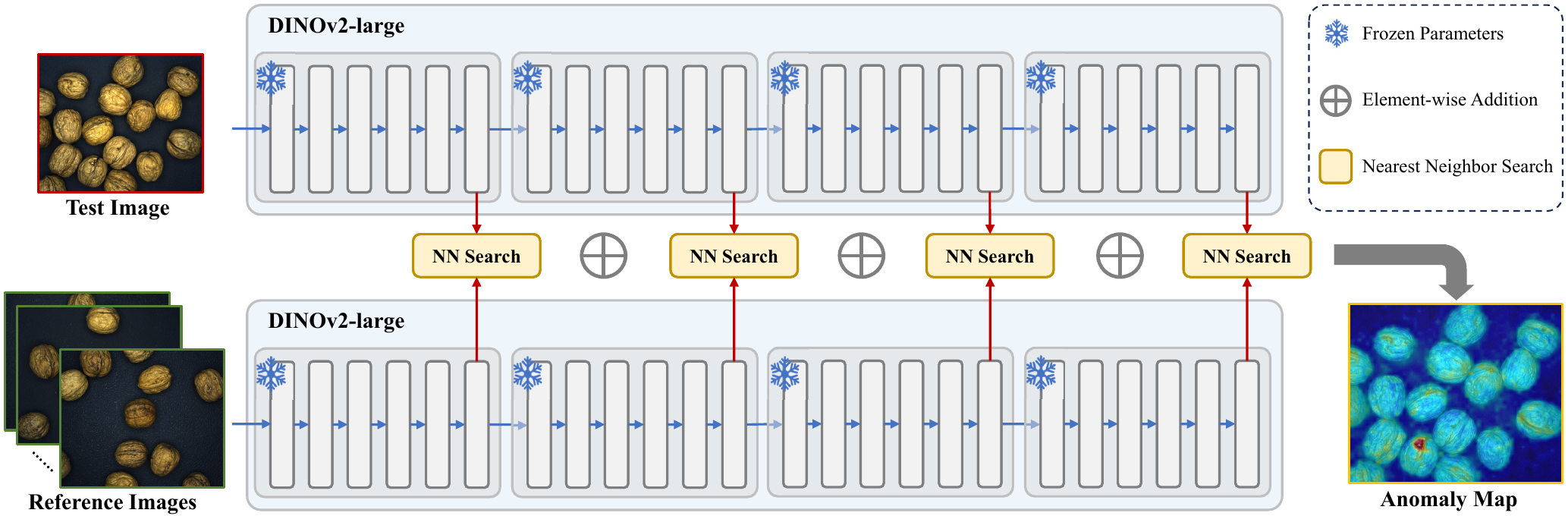}
  \caption{The overall architecture of our proposed method.}
  \label{fig:architecture}
\end{figure*}

\subsubsection{Architecture}
\label{sec:Architecture}
The overall architecture of our proposed method is illustrated in Figure~\ref{fig:architecture}.

For each category in the dataset, we construct a memory bank consisting of 16 normal reference images. The selection of these reference images follows a two-step procedure. First, we extract CLS token representations from all training images using the DINOv2 model. Then, the same greedy coreset selection method as used in Patchcore is applied to group these feature vectors into 16 clusters. This strategy maximizes the coverage of diverse normal patterns within a category without changing the distribution of all features, thereby enhancing the representativeness of the reference set and reducing false positive rates.

For each test image, we first extract multi-level features using the DINOv2 model. For each level of features, we compute the similarity between the test image features and those stored in the memory bank to retrieve the nearest neighbor information. Our method is based on the assumption that normal regions in the test image tend to find similar regions among the reference images, while anomalous regions lack such matches. By evaluating the similarity at each spatial location, we obtain an anomaly map for each level. Finally, these anomaly maps are averaged and upsampled to the original resolution to generate the final anomaly segmentation map.

\subsubsection{Training}
Our proposed method requires no training. For each category in the MVTec AD 2 dataset, we construct a few-shot feature memory bank of normal reference images using samples from the training set. The number of reference samples is fixed at 16. We adopt the pretrained DINOv2-ViT-L-14 model as the feature extractor, which consists of 24 transformer layers and approximately 300 million parameters. Features are extracted from four specific layers (i.e., layer 6, 12, 18, and 24) to generate the final anomaly segmentation map.

To ensure memory efficiency, input image resolutions are adjusted: for all categories except \texttt{Sheet Metal}, the shorter side is resized to 672 pixels while preserving the original aspect ratio. For \texttt{Sheet Metal}, due to the elongated image size, the shorter side is resized to 448 pixels, also preserving the original aspect ratio. This configuration allows our method to run entirely on a single 24GB GPU (e.g., NVIDIA GeForce RTX 3090).

For the \texttt{Vial} and \texttt{Wallplugs} categories, we further enhance segmentation performance by extracting foreground features from the input images, as detailed in Section~\ref{sec:Dataset Utilization}.

For the \texttt{Fabric} and \texttt{Walnuts} categories, there are cases where the central pattern is entirely normal while anomalies exist only at the edges (e.g., in the \texttt{Fabric} category, a piece of fabric may be placed over the background fabric). After the initial detection, we apply post-processing to these two categories by filling the interiors of closed regions, further improving prediction accuracy.

\subsection{Dataset \& Evaluation}
\subsubsection{Dataset Utilization}
\label{sec:Dataset Utilization}
Since our proposed method is training-free, for each category, we only utilize images from its training set to construct a few-shot memory bank. In this method, we innovatively propose an adaptive background mask generation technique based on Principal Component Analysis (PCA) and morphological optimization for the image preprocessing stage. This enhances visual feature representation while suppressing redundant background information.

First, we apply PCA to reduce the dimensionality of features extracted from the input image and extract the first principal component. This technique captures the direction of maximum variance in the feature space by performing singular value decomposition on the covariance matrix, which is mathematically expressed as:

\begin{equation}
\mathbf{PC}_1 = \arg\max_{\|\mathbf{v}\| = 1} \mathrm{Var}(\mathbf{Xv}),
\end{equation}
where $\mathbf{X}$ denotes the normalized feature matrix, and $\mathbf{v}$ is the projection vector. The first principal component reflects the primary mode of variation in the feature distribution. Subsequently, we binarize the projection values based on a predefined threshold $\tau$ to generate an initial mask:

\begin{equation}
\mathcal{M}_{\text{init}}[i] = 
\begin{cases}
1 & \text{if } \mathbf{PC}_1[i] > \tau \\
0 & \text{otherwise}
\end{cases}.
\end{equation}

Since it is initially unclear which part corresponds to foreground or background, we propose an adaptive decision strategy based on variance analysis. Specifically, we initially designate one region as foreground and the other as background, then compute the feature variance within each region and compare their medians. If the variance of the foreground region is lower than that of the background region, it suggests a potential misclassification, and we invert the mask accordingly. This strategy effectively mitigates foreground-background misclassification caused by improper thresholding.

\begin{equation}
\small
\mathcal{M}_{\text{init}} = 
\begin{cases}
\mathcal{M}_{\text{init}} & \text{if } \mathrm{MD}(\mathrm{Var}(\mathbf{F}_{\text{msk}})) \geq \mathrm{MD}(\mathrm{Var}(\mathbf{F}_{\text{n-msk}})) \\
\neg \mathcal{M}_{\text{init}} & \text{otherwise}
\end{cases},
\end{equation}
where $M_{\text{init}}$ denote the initial binary mask (1 for masked region, 0 for unmasked region), $F_{\text{msk}} \in \mathbb{R}^{N_{\text{msk}} \times D}$ and $F_{\text{n-msk}} \in \mathbb{R}^{N_{\text{n-msk}} \times D}$ represent the feature matrices of the masked and unmasked regions, respectively. $\text{Var}(\cdot)$ computes the per-channel variance (dimension: $D \times 1$), and $\mathrm{MD}(\cdot)$ returns the median value of the variance vector to suppress the influence of outliers. The operator $\lnot$ denotes boolean mask inversion.

To address discrete noise in the initial mask, we apply 2D morphological operations for post-processing. Specifically, we use a $k \times k$ square kernel to dilate the mask, enhancing region connectivity. By combining dilation and erosion, we eliminate holes and smooth the boundaries. This process is formally defined as:

\begin{equation}
\mathcal{M}_{\text {final }}=\operatorname{Closing}\left(\operatorname{Dil.}\left(\mathcal{M}_{\text {init }} \in\{0,1\}^{H \times W}\right), \mathcal{K}\right),
\end{equation}
where $\mathcal{K}$ is the morphological kernel and $H \times W$ is the spatial dimension of the feature grid. Finally, the optimized 2D mask is reshaped back to the feature vector dimension, yielding a boolean mask matrix used for element-wise filtering of the original features. This effectively suppresses interference from low-information background regions.

In all experiments, we set the number of PCA components to 1, the threshold $\tau$ to 1.0, and the morphological kernel size to $3 \times 3$. These parameters generalize well across various datasets and require no dataset-specific tuning. Experimental results demonstrate that the proposed preprocessing method preserves the integrity of the main object features while significantly reducing background noise interference, thereby improving model performance.

\begin{table}[t]
\centering
\begin{tabular}{l c c}
\toprule
Object & AU-ROC\textsubscript{0.05} & F1 Score \\
\midrule
Can & 58.61 & 0.18 \\
Fabric & 68.29 & 28.22 \\
Fruit Jelly & 80.93 & 48.27 \\
Rice & 92.92 & 68.44 \\
Vial & 69.04 & 35.88 \\
Wallplugs & 77.90 & 19.19 \\
Walnuts & 89.23 & 75.05 \\
Sheet Metal & 76.80 & 40.13 \\
\midrule
Mean & 76.71 & 39.42 \\

\bottomrule
\end{tabular}
\caption{AU-ROC\textsubscript{0.05} and segmentation F1 score (in\%) on binarized images for \textit{TEST\textsubscript{public}} set.}
\label{tab:result_public}
\end{table}

\begin{table*}[t]
\centering
\resizebox{\textwidth}{!}{
\begin{tabular}{lcccccccc}
\toprule
Object & PatchCore~\cite{roth2022towards} & RD~\cite{deng2022anomaly} & RD+~\cite{tien2023revisiting} & EfficientAD~\cite{batzner2024efficientad} & MSFlow~\cite{zhou2024msflow} & SimpleNet~\cite{liu2023simplenet} & DSR~\cite{zavrtanik2022dsr} & Ours\\
\midrule
Can & 0.3 / 0.1 & 0.1 / 0.1 & 0.1 / 0.1 & 0.8 / 0.1 & 5.0 / 0.1 & 0.6 / 0.1 & 0.4 / 0.1 & \textbf{17.3} / \textbf{1.9}\\
Fabric & 11.5 / 9.8 & 2.6 / 2.2 & 2.9 / 2.3 & 7.6 / 1.0 & 22.0 / 4.1 & 21.6 / 10.2 & 7.9 / 5.0 & \textbf{77.4} / \textbf{65.3} \\
Fruit Jelly & 8.7 / 8.2 & 22.5 / 22.7 & 26.9 / 26.7 & 20.8 / 18.2 & \textbf{47.6} / 38.1 & 25.1 / 23.0 & 17.9 / 17.2 & 41.3 / \textbf{40.9} \\
Rice & 3.8 / 4.2 & 7.0 / 3.9 & 9.5 / 2.9 & 15.0 / 0.5 & 19.1 / 1.8 & 11.6 / 1.0 & 1.5 / 1.4 & \textbf{60.9} / \textbf{61.2}\\
Sheet Metal & 1.8 / 1.1 & 41.3 / 39.2 & 40.9 / 37.7 & 9.3 / 3.8 & 13.0 / 7.6 & 14.6 / 2.8 & 13.9 / 14.4 & \textbf{59.5} / \textbf{59.7} \\
Vial & 2.3 / 2.2 & 28.0 / 28.3 & 28.2 / 22.8 & 30.5 / 26.5 & 23.3 / 6.2 & 31.9 / 17.5 & 28.2 / 27.9 & \textbf{42.8} / \textbf{40.8} \\
Wallplugs & 0.0 / 0.0 & 1.9 / 0.8 & 1.3 / 0.9 & 4.4 / 0.3 & 0.1 / 0.2 & 1.0 / 0.3 & 0.4 / 0.4 &  \textbf{13.7} / \textbf{6.7}\\
Walnuts & 1.2 / 1.3 & 41.2 / 36.7 & 44.1 / 40.5 & 34.6 / 13.3 & 44.5 / 14.3 & 35.2 / 14.3 & 17.0 / 9.6 & \textbf{69.1} / \textbf{69.1} \\
\midrule
Mean & 3.7 / 3.4 & 18.1 / 16.7 & 19.2 / 16.7 & 15.4 / 8.0 & 21.8 / 9.0 & 17.7 / 8.7 & 10.9 / 9.5 & \textbf{47.8} / \textbf{43.2}\\
\bottomrule
\end{tabular}
}
\caption{Performance comparison of segmentation F1 score (in\%) on binarized images for \textit{TEST\textsubscript{priv}} / \textit{TEST\textsubscript{priv,mix}} set. The best results are highlighted in bold.}
\label{tab:comparison}
\end{table*}

\subsubsection{Evaluation Criteria}
We evaluate the model's performance primarily using the \textbf{pixel-level F1 score}. This metric combines precision and recall by computing their harmonic mean, thereby providing a balanced measure of the model’s ability to detect anomalies at the pixel level. Specifically, the F1 score is calculated as:

\begin{equation}
\text{F1} = 2 \times \frac{\text{Precision} \times \text{Recall}}{\text{Precision} + \text{Recall}},
\end{equation}
where \text{Precision} denotes the proportion of predicted anomalous pixels that are truly anomalous, while \text{Recall} represents the proportion of actual anomalous pixels that are correctly identified by the model.

During evaluation, we optimize the F1 score by adjusting the decision threshold on \textit{TEST\textsubscript{public}} dataset to determine the optimal boundary for anomaly segmentation. This process ensures that the model achieves a balanced trade-off between precision and recall.

The use of the pixel-level F1 score as the evaluation metric enables precise assessment of the model's capability to identify anomalous regions in complex image data. Notably, the MVTec AD 2 dataset emphasizes the detection of \textbf{small defects}, which may occupy only a few pixels in an image. In such cases, conventional metrics like the Area Under the Receiver Operating Characteristic Curve (AU-ROC) can be dominated by larger defects, thus failing to accurately reflect the model's performance on smaller anomalies. In contrast, the pixel-level F1 score places equal emphasis on detecting all anomalous regions, regardless of their size. This makes it particularly well-suited to the challenges posed by the MVTec AD 2 dataset, as it can reliably evaluate the model’s ability to correctly identify even the smallest defects.

%% file: sec/3_results.tex
\section{Results}

\subsection{Performance Metrics}
Our proposed method demonstrates excellent performance in pixel-level anomaly detection tasks. Specifically, on the \textit{TEST\textsubscript{public}} dataset, the model's performance in terms of AU-ROC\textsubscript{0.05} and segmentation F1 score is summarized in Table~\ref{tab:result_public}.

Moreover, based on the official test results from the VAND 3.0 Challenge server, our model achieves an F1 score of \textbf{47.18\%} on the \textit{TEST\textsubscript{priv}} dataset and \textbf{42.51\%} on the \textit{TEST\textsubscript{priv,mix}} dataset. These results indicate that the model achieves a good balance between precision and recall, and can effectively detect anomalous pixels in images. Notably, the model maintains relatively stable performance even under distributional shifts in the data.

To more comprehensively evaluate the model’s performance, we also consider other key metrics. On the \textit{TEST\textsubscript{priv}} dataset, the model achieves an AucPro\textsubscript{0.05} score of \textbf{60.51\%}, while on the \textit{TEST\textsubscript{priv,mix}} dataset, the score is \textbf{58.37\%}. These results suggest that the model has strong detection capabilities for anomalous regions across different thresholds and remains accurate even under variations in illumination and other environmental factors.

In addition, for image-level classification tasks, the model achieved ClassF1 scores of \textbf{70.2\%} and \textbf{74.4\%} on the \textit{TEST\textsubscript{priv}} and \textit{TEST\textsubscript{priv,mix}} datasets, respectively, demonstrating strong capability in distinguishing between normal and anomalous samples at the image level.

\subsection{Comparison}
Due to the recent release of the MVTec AD 2 dataset and the unavailability of Ground Truth for the \textit{TEST\textsubscript{priv}} and \textit{TEST\textsubscript{priv,mix}} test sets, we compare our proposed method with other approaches listed in the MVTec AD 2 dataset paper, as shown in Table~\ref{tab:comparison}. On both test sets, our method, SuperAD, consistently outperforms previous methods. Moreover, our method requires no training, demonstrating superior generalization capabilities compared to the other approaches.

%% file: sec/4_discussion.tex
\section{Discussion}

\subsection{Challenges \& Solutions}
During our experiments, we observe that certain categories (such as \texttt{Fruit Jelly}, \texttt{Vial} and \texttt{Wallplugs}) exhibit high intra-class variability. When constructing the reference feature memory bank using randomly selected images for these categories, many normal regions are incorrectly identified as anomalies due to the lack of sufficiently similar patterns in the memory bank. To address this issue, we propose selecting reference images using a greedy coreset selection strategy rather than random sampling. This approach increases the diversity of patterns within the memory bank and helps reduce false positives. A detailed explanation of this method is provided in Section~\ref{sec:Architecture}.

We further analyze the anomaly segmentation maps generated from the similarity scores between the extracted features at four different layers and the reference images for each category. We observe that, due to the relatively clear separation between foreground and background in most categories of the MVTec AD 2 dataset, false positives in background regions are generally limited. However, in the case of the \texttt{Wallplugs} category, some false detections still occur in the background owing to its highly complex and diverse patterns.

To gain deeper insights, we evaluate the effectiveness of PCA-based binary classification on the features extracted by DINOv2. Our results indicate that applying PCA to the shallow layers of DINOv2, with a threshold set to 1, effectively separates foreground and background regions. More details can be found in Section~\ref{sec:Dataset Utilization}.

Based on this observation, we apply a foreground feature extraction preprocessing strategy to the \texttt{Wallplugs} category, which leads to improved segmentation performance for this particularly challenging class.

\subsection{Model Robustness \& Adaptability}
It is worth emphasizing that our proposed method is entirely training-free, and thus does not require fine-tuning DINOv2 for any specific category. As a result, the generalization ability of the model remains fully preserved. 

Our approach relies heavily on the powerful feature extraction capabilities of DINOv2, which enables the extraction of semantically rich representations suitable for comparison across diverse categories. 

During the construction of the memory bank, we not only select 16 reference images but also employ a greedy coreset selection strategy to ensure that the selected images are as diverse as possible. This strategy effectively alleviates the issue of limited pattern diversity within the memory bank, which might otherwise fail to represent the full range of intra-class variability. Consequently, this design further enhances the robustness of our model.

\subsection{Future Work}
Although our method achieves competitive performance, several prominent issues remain unresolved and warrant further investigation.

\begin{figure}[t]
  \centering
  \includegraphics[width=1.0\linewidth]{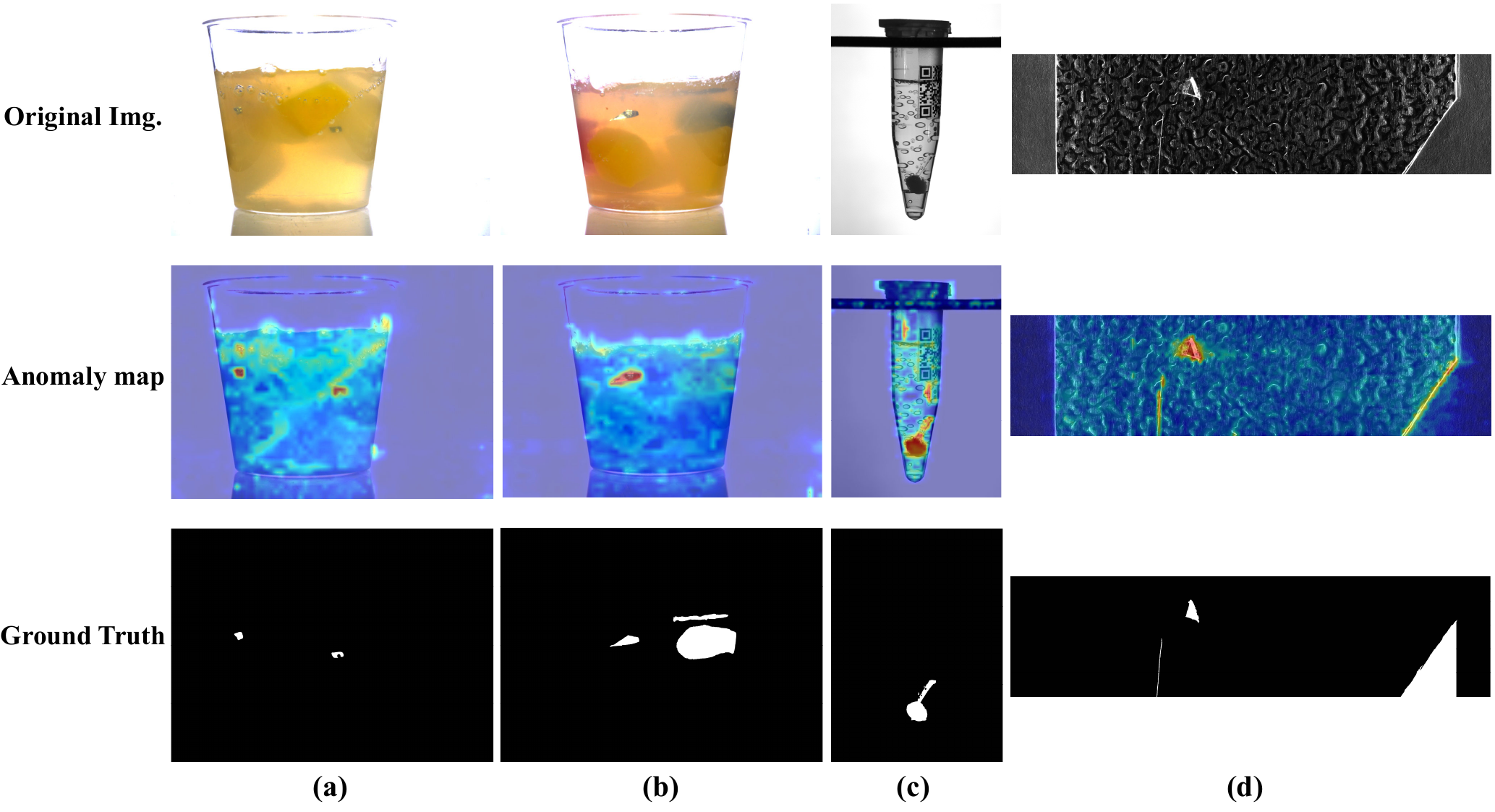}
  \caption{Examples of typical failure cases: (a) false positives on air bubbles; (b) missed detection of blurred objects; (c) false positives due to specular highlights; (d) missed detections of missing-type anomalies.}
  \label{fig:limitations}
\end{figure}

\textbf{False positives on air bubbles:} As illustrated in Figure~\ref{fig:limitations}(a), \texttt{Fruit Jelly} category contains a wide variety of air bubbles with high uncertainty. These false positives are primarily caused by the diversity of bubble appearances, which often do not match the bubble patterns in the fixed memory bank of normal features. In the future, the robustness of the model to complex pattern variations could be enhanced to reduce such misclassifications.

\textbf{Missed detections of reflections and blurry objects:} As shown in Figure~\ref{fig:limitations}(b), in \texttt{Fruit Jelly} category, light-colored reflections and light scattering caused by the optical properties of jelly can diminish the perceived abnormality of dark objects. The model may erroneously match such regions to normal patterns. Future work could improve the model's ability to recognize objects within transparent or semi-transparent media to reduce missed detections.

\textbf{False positives on specular highlights:} As depicted in Figure~\ref{fig:limitations}(c), in \texttt{Vial} and \texttt{Can} categories, specular highlights caused by illumination often lead to false predictions. Incorporating advanced illumination compensation or highlight removal techniques may enhance the model's robustness to such lighting artifacts and reduce false positives.

\textbf{Missed detections of missing-type anomalies:} As shown in Figure~\ref{fig:limitations}(d), in \texttt{Sheet Metal} and \texttt{Vial} categories, missing-type anomalies are sometimes difficult to detect due to their high visual similarity with the background. Future research may explore ways to enable model to better capture object completeness features, thereby improving its ability to detect missing-type anomalies.

%% file: sec/5_conclusion.tex
\section{Conclusion}
In this report, we propose a fully training-free anomaly detection and segmentation method that achieves robust performance under complex and varying lighting conditions. In our method, we first employ a greedy coreset selection strategy to select a small number of diverse normal reference images. Then, leveraging the powerful representational capacity of the DINOv2 model, we extract image features from the selected references to construct a memory bank. For a test image, we extract multi-scale features and perform nearest neighbor matching with the memory bank at each scale to generate anomaly segmentation maps. These maps are then averaged to produce the final result. 

Our method demonstrates that the large pre-trained model DINOv2 possesses excellent image representation capabilities. Relying solely on these capabilities without any fine-tuning enables effective performance on dense prediction tasks such as anomaly segmentation.